\documentclass[letterpaper, 10 pt, conference]{ieeeconf}  

\IEEEoverridecommandlockouts                              

\usepackage{amsmath} 
\usepackage{amssymb}  

\usepackage{tikz}
\usetikzlibrary{calc}
\usepackage{xcolor}
\usepackage{bm}
\usepackage{hyperref}
\usepackage{booktabs}
\usepackage{makecell}
\usepackage{siunitx}

\definecolor{TUMBlau}{RGB}{0,101,189} 
\definecolor{TUMBlauDunkel}{RGB}{0,82,147} 
\definecolor{TUMBlauHell}{RGB}{152,198,234} 
\definecolor{TUMBlauMittel}{RGB}{100,160,200} 

\definecolor{TUMElfenbein}{RGB}{218,215,203} 
\definecolor{TUMGruen}{RGB}{162,173,0} 
\definecolor{TUMOrange}{RGB}{227,114,34} 
\definecolor{TUMGrau}{gray}{0.6} 

\title{\LARGE \bf
Rolling-Shutter Modelling for Direct Visual-Inertial Odometry}

\author{David Schubert$^{1,2}$, Nikolaus Demmel$^{1}$,  Lukas von Stumberg$^{1,2}$, Vladyslav Usenko$^{1}$ and Daniel Cremers$^{1,2}$
\thanks{$^{1}$The authors are with the Computer Vision Group, Technical University of Munich, Germany,
        {\tt\small \{schubdav, demmeln, stumberg, usenko, cremers\}@in.tum.de}}%
\thanks{$^{2}$The authors are with Artisense Corporation}%
}

\DeclareMathOperator{\SE}{SE}
\DeclareMathOperator{\se}{\mathfrak{se}}
\DeclareMathOperator{\SO}{SO}

\DeclareMathOperator{\obs}{obs}
\DeclareMathOperator{\Adj}{Adj}
\newcommand{\mat}[1]{\mathbf{#1}}
\newcommand{\matsym}[1]{\mathbf{#1}}
\renewcommand{\vec}[1]{\mathbf{#1}}
\newcommand{\vecsym}[1]{\bm{#1}}
\newcommand{\R}{\mathbb{R}}
\usepackage{siunitx}

\begin{document}

\maketitle
\thispagestyle{empty}
\pagestyle{empty}

\begin{abstract}

We present a direct visual-inertial odometry (VIO) method which estimates the motion of the sensor setup and sparse 3D geometry of the environment based on measurements from a rolling-shutter camera and an inertial measurement unit (IMU).

The visual part of the system performs a photometric bundle adjustment on a sparse set of points. This direct approach does not extract feature points and is able to track not only corners, but any pixels with sufficient gradient magnitude. Neglecting rolling-shutter effects in the visual part severely degrades accuracy and robustness of the system. In this paper, we incorporate a rolling-shutter model into the photometric bundle adjustment that estimates a set of recent keyframe poses and the inverse depth of a sparse set of points.

IMU information is accumulated between several frames using measurement preintegration, and is inserted into the optimization as an additional constraint between selected keyframes. For every keyframe we estimate not only the pose but also velocity and biases to correct the IMU measurements. Unlike systems with global-shutter cameras, we use both IMU measurements and rolling-shutter effects of the camera to estimate velocity and biases for every state.

Last, we evaluate our system on a new dataset that contains global-shutter and rolling-shutter images, IMU data and ground-truth poses for ten different sequences, which we make publicly available. Evaluation shows that the proposed method outperforms a system where rolling shutter is not modelled and achieves similar accuracy to the global-shutter method on global-shutter data.

\end{abstract}

\section{INTRODUCTION}

Many robotics applications rely on motion estimation and 3D reconstruction. Laser rangefinders, RGB-D cameras \cite{kerl13icra}, GPS and many other sensors can be used to solve these tasks, but cameras are the most popular choice for such applications, because they are cheap, lightweight and small. They are passive sensors, so they do not interfere with each other when placed in the same environment. Several works have shown the application of cameras for robot navigation 
\cite{weiss12realtime, stelzer12ijrr} and autonomous driving \cite{geiger12cvpr}.

\begin{figure}[t]
    \centering
    \includegraphics[width=\linewidth]{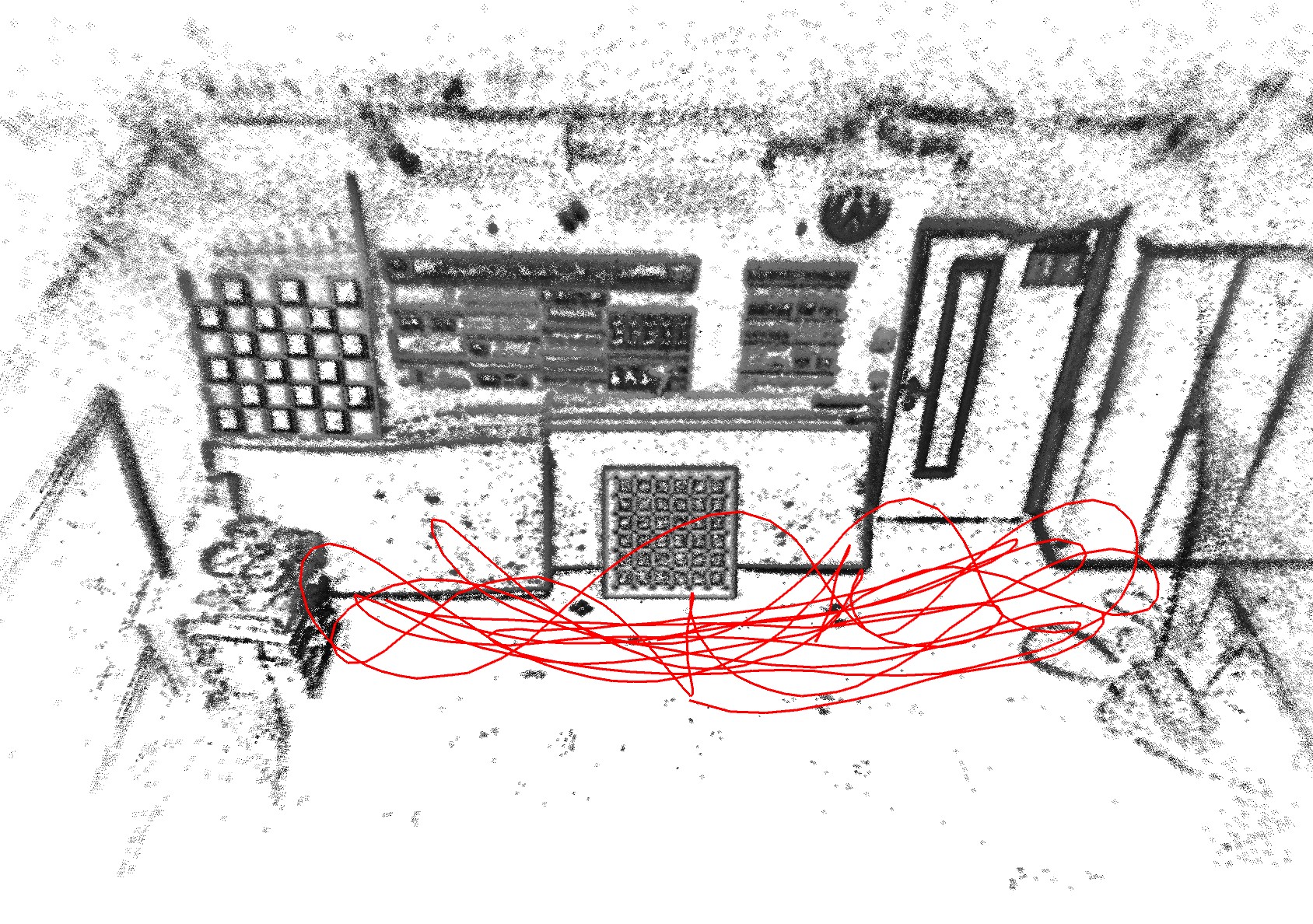}
    \caption{Reconstructed camera trajectory (red) and sparse 3D reconstruction of our method on sequence 6 of our new dataset. Despite diverse motion patterns that revisit different parts of the scene multiple times, edges in the sparse point cloud stay very consistent and show little drift.}
    \label{fig:qualitative}
\end{figure}

\begin{figure}[t]
    \centering
    \includegraphics[width=0.48\linewidth]{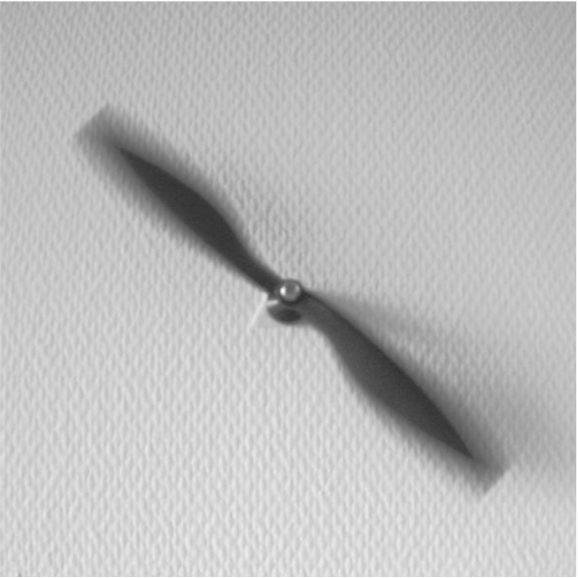}
    \includegraphics[width=0.48\linewidth]{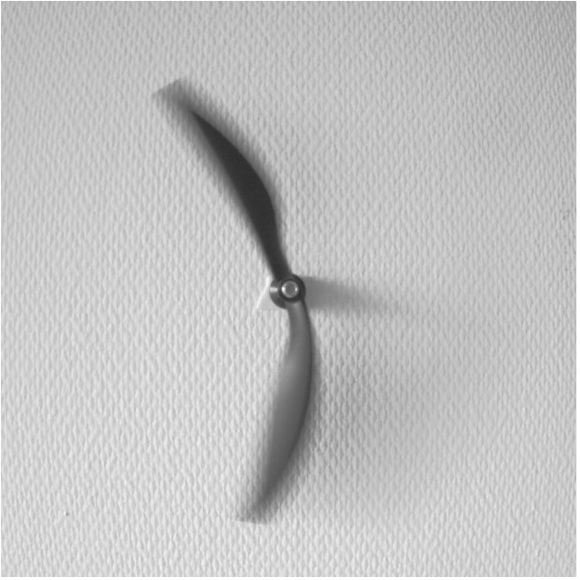}
    \caption{Difference between global-shutter (left) and rolling-shutter (right) images when exposed to fast motion. Both images were triggered at the same time. The rotating propeller appears distorted when a rolling-shutter sensor is used.}
    \label{fig:gs_vs_rs}
\end{figure}

Most visual odometry methods assume that cameras have a global shutter, and with this assumption show impressive results in 3D reconstruction and motion estimation (e.g.~\cite{engel2018direct,mur2017orb}).
A global-shutter camera exposes all pixels in the image simultaneously. However, rolling-shutter CMOS sensors are widespread in consumer devices (e.g.~tablets, smartphones), but also in the automotive sector and in the motion picture industry. A rolling-shutter camera exposes rows sequentially with some delay and reads them one by one. This leads to large image distortions in the presence of fast motion, as can be seen in Fig.~\ref{fig:gs_vs_rs}. Neglecting rolling-shutter effects leads to significant drift in the estimated trajectory and inaccurate 3D reconstruction~\cite{schubert2018direct}, but when these effects are modelled correctly the system can achieve accuracy similar to global-shutter systems (Fig.~\ref{fig:qualitative}).

There exist two major types of approaches for visual odometry. Indirect methods (e.g.~\cite{mur2017orb})
align pixel coordinates of the matched keypoints, whereas direct methods (e.g.~\cite{engel2018direct}) align image intensities based on the photoconsistency assumption. Direct methods outperform indirect methods in weakly textured environments, but they are more sensitive to geometric noise~\cite{engel2018direct}. Rolling-shutter effects introduce strong geometric changes and thus, for direct methods, it is much more important to model rolling shutter to achieve good results than for indirect methods. Unlike indirect methods, with direct methods the capture time of a point in its target frame is not directly known after selecting the point in its host frame, so the \emph{rolling-shutter constraint} \cite{meingast2005geometric} has to be imposed in order to retrieve the capture time.

Another challenge for visual odometry methods is the lack of robustness in areas with low number of features, or when performing fast maneuvers. In the case of monocular cameras they are also not able to reconstruct the scale of the environment. By combining a camera with an inertial measurement unit (IMU) we can make the system more robust and, given sufficient excitation, estimate the metric scale of the environment. IMU measurements do not suffer from outliers and with corrected bias provide accurate short-term motion prediction.

In this paper, we propose a novel direct visual-inertial odometry method for rolling-shutter cameras.
Our approach estimates pose, linear velocity and biases for each keyframe and the inverse depth of the points that are tracked by the system. To model the continuous motion of the camera we also optimize a twist in the camera frame that is coupled to the IMU velocity and biases, and use a constant-twist motion assumption to represent the continuous trajectory. This way we can incorporate rolling-shutter effects into the optimization.

We evaluate our method on ten challenging sequences from a newly recorded dataset that we make publicly available. The dataset features not only IMU data and rolling-shutter images, but also simultaneously recorded global-shutter images for comparison. To our knowledge, such a dataset currently does not exist. We compare our method to a state-of-the-art approach for global-shutter cameras, demonstrating that systems that model rolling shutter can achieve similar performance to global-shutter VIO methods running on global-shutter data.

In summary, our contributions are:
\begin{itemize}
    \item a tight integration of rolling-shutter visual and inertial information in a direct odometry system,
    \item velocity and bias estimation not only from the IMU measurements, as in other methods for global-shutter visual-intertial odometry, but also from rolling-shutter effects of the images,
    \item a dataset that contains sequences simultaneously captured with global-shutter and rolling-shutter cameras that are time-aligned with IMU and motion capture data,
    \item an extensive evaluation of the proposed system on the collected dataset and comparison to the baseline global-shutter method.
\end{itemize}

The dataset and additional information about the system are available on:
\begin{center}
\color{purple}
\urlstyle{tt}
\textbf{\href{https://vision.in.tum.de/data/datasets/rolling-shutter-dataset}{\nolinkurl{https://vision.in.tum.de/data/datasets/}}\\
\href{https://vision.in.tum.de/data/datasets/rolling-shutter-dataset}{\nolinkurl{rolling-shutter-dataset}}}
\end{center}

\section{RELATED WORK}

Visual-inertial odometry (VIO) can be grouped into two major approaches. Filtering-based methods keep a probabilistic representation of the state of the system, that includes a mean and a covariance matrix to capture correlations between variables. One example is ROVIO~\cite{Bloesch, bloesch2017iterated}, that uses an Extended Kalman Filter (EKF) and photometric residuals by comparing image patches, which are tightly coupled. An important extension of the EKF is the Multi-State Constraint Kalman Filter (MSCKF)~\cite{mourikis2007multi, mourikis13IJRR}, that includes constraints from observing features in multiple images and does not require feature positions in the state vector, which yields a computational complexity linear in the number of features. This has already led to a variant for the rolling-shutter case~\cite{li2013real}.

On the other hand, optimization-based approaches have gained popularity. They set up an energy function that incorporates models of the sensors and perform a non-linear optimization to estimate parameters. Information from older frames can be kept in the system using marginalization. This approach has been successfully demonstrated with OKVIS~\cite{leutenegger2014keyframe, leutenegger2013keyframe}. Direct examples of optimization-based VIO are given in~\cite{usenko2016direct, von2018direct}. The latter is based on DSO~\cite{engel2018direct}, a state-of-the-art monocular visual odometry system.
In its optimization backend, a global bundle adjustment is performed on a set of recent keyframes and a sparse set of points.
In~\cite{von2018direct}, the method is extended with a tightly coupled IMU integration and a method to tackle delayed scale observability in the presence of marginalization priors, which is a problem when the scale moves too far from its linearization point. Their strategy is to keep two marginalization priors with different linearization points and switch to the newer one when needed. A full visual-inertial SLAM system is given by VINS-Mono~\cite{qin2018vins}, which is also based on non-linear optimization. A visual-inertial extension of ORB-SLAM~\cite{mur2015orb, mur2017orb}, a state-of-the-art keyframe-based SLAM approach, is given in~\cite{mur2017visual}. Another method~\cite{lovegrove2013spline} proposes a B-spline representation of the trajectory to incorporate rolling shutter and measurements of different sensors.

Modelling rolling shutter in the domain of direct odometry methods has been attempted with different sensor modalities. The RGB-D method in~\cite{kerl13icra} uses not only a photometric error term, but also a geometric error term due to the availability of depth measurements. Direct monocular approaches have been presented in \cite{kim2016direct,schubert2018direct}. While the first decouples motion and structure estimation and enforces the rolling-shutter constraint softly by introducing additional time variables, the latter performs a global bundle adjustment and explicitly solves the rolling-shutter constraint.

Contrary to the other methods, we present a direct visual-inertial rolling-shutter odometry method, which combines existing rolling-shutter~\cite{schubert2018direct} and visual-inertial~\cite{von2018direct} approaches and couples the underlying variables with a new energy term.

\section{NOTATION}

In this paper, poses from the special Euclidean group $\SE(3)$ are represented as $4\times 4$ matrices
\begin{align}
\mat{T}=
    \begin{bmatrix}
    \mat{R} & \vec{t} \\
    \bm{0}_{1\times 3} & 1
    \end{bmatrix}\,,
\end{align}
where $\mat{R}\in\SO(3)$ is a rotation matrix from the special orthogonal group, and $\vec{t}\in\R^3$ is a three-dimensional translation component.

In order to optimize scale and gravity direction, we will also use transformations from $\R^+\times\SO(3)$, which include a positive scale $s\in\R^+$ and a rotation matrix $\mat{R}\in\SO(3)$ and act as the matrix
\begin{align}
\mat{T} =
\begin{bmatrix}
    s\mat{R} & \bm{0}_{3\times 1} \\
    \bm{0}_{1\times 3} & 1
    \end{bmatrix}\,,
\end{align}
so they can also be seen as a similarity transform with zero translation component.

We will also make use of the exponential map,
\begin{align}
    \exp\colon \se(3)\rightarrow\SE(3)\,,
\end{align}
to map elements from the Lie algebra $\se(3)$ to the Lie group $\SE(3)$, which, in matrix representation, is just the matrix exponential (but has a closed form in this particular case). Lie algebra elements $\hat{\vecsym{\xi}}$ are generated from vectors $\vecsym{\xi}\in\R^6$ using the hat operator. We use the convention that the first three components of $\vecsym{\xi}$ correspond to translation and the last three components correspond to rotation. Using the exponential map, it is possible to parametrize poses as a function of time as
\begin{align}
    \mat{T}(t)=\exp(\hat{\vecsym{\xi}}t)\mat{T}_0\,,
\end{align}
starting from pose $\mat{T}_0\in\SE(3)$ at $t=0$.
This is called a constant-twist interpolation. For brevity, we will not use the hat operator inside the exponential function. Whenever the exponential function acts on a vector, it denotes a composition of the hat operator and the exponential. Similarly, we will call 6D vectors twists.

When representing 3D points in different coordinate systems $A$ and $B$, the pose that converts a point's representation $\vec{p}_A$ in system $A$ to its representation $\vec{p}_B$ in system $B$ is written $\mat{T}_{BA}$, and the transformed point is calculated as
\begin{align}
    \vec{p}_B = \mat{T}_{BA}\vec{p}_A\,,
\end{align}
where 3D points $(x,y,z)^\top$ are represented as $\vec{p}=(x,y,z,1)^\top$.

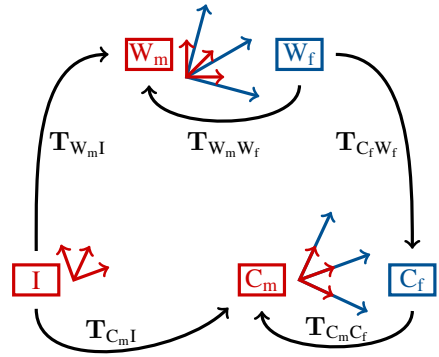
\begin{figure}

\centering

\begin{tikzpicture}

\definecolor{airforceblue}{rgb}{0.36, 0.54, 0.66}
\definecolor{bostonuniversityred}{rgb}{0.8, 0.0, 0.0}

\newcommand{\colorScale}{bostonuniversityred}
\newcommand{\colorFree}{TUMBlauDunkel}
\newcommand{\textScale}[1]{{\color{\colorScale}#1}}
\newcommand{\textFree}[1]{{\color{\colorFree}#1}}
\newcommand{\cssymbol}[1]{
	\draw[very thick,<->,#1] (0.5,0)--(0,0) -- (0,0.5);
	\draw[very thick,->,#1] (0,0) -- (45:0.5);
}
\newcommand{\cslabel}[3]{
    \begin{scope}[#2, local bounding box=#3]
        \draw[very thick] (-0.3,-0.2) rectangle (0.3,0.2);
        \node {#1};
    \end{scope}
}

\cssymbol{\colorScale, rotate=20}
\cssymbol{\colorFree, xshift=3cm, scale=2, rotate=-25}
\cssymbol{\colorScale, xshift=3cm, rotate=-25}
\cssymbol{\colorFree, xshift=1.5cm,yshift=2.7cm, scale=2, rotate=-15}
\cssymbol{\colorScale, xshift=1.5cm,yshift=2.7cm}

\cslabel{$\text{I}$}{\colorScale, xshift = -0.5cm}{i}
\cslabel{$\text{C}_\text{m}$}{\colorScale, xshift = 2.5cm}{cs}
\cslabel{$\text{C}_\text{f}$}{\colorFree, xshift = 4.5cm}{cf}
\cslabel{$\text{W}_\text{m}$}{\colorScale, xshift = 1cm, yshift=3cm}{ws}
\cslabel{$\text{W}_\text{f}$}{\colorFree, xshift = 3cm, yshift=3cm}{wf}

\draw[very thick,->, shorten >=0.15cm, shorten <=0.15cm] (i.north) to [out=90, in=180] (ws.west);
\draw[very thick,<-, shorten >=0.15cm, shorten <=0.15cm] (ws.south) to [out=-90, in=-90] (wf.south);
\draw[very thick,->, shorten >=0.15cm, shorten <=0.15cm] (wf.east) to [out=0, in=90] (cf.north);
\draw[very thick,<-, shorten >=0.15cm, shorten <=0.15cm] (cs.south) to [out=-90, in=-90] (cf.south);
\draw[very thick,->, shorten >=0.15cm, shorten <=0.15cm] (i.south) to [out=-90, in=-135] (cs.south west);

\node at (0.5, -0.7) {$\mat{T}_{\text{C}_\text{m}\text{I}}$};
\node at (3.5, -0.65) {$\mat{T}_{\text{C}_\text{m}\text{C}_\text{f}}$};
\node at (3.9, 1.8) {$\mat{T}_{\text{C}_\text{f}\text{W}_\text{f}}$};
\node at (0.05, 1.8) {$\mat{T}_{\text{W}_\text{m}\text{I}}$};
\node at (2, 1.8) {$\mat{T}_{\text{W}_\text{m}\text{W}_\text{f}}$};

\end{tikzpicture}

\caption{Coordinate systems and transformations used in this paper. The coordinate system abbreviations are: $\text{I}$: IMU, $\text{W}$: world, $\text{C}$: camera. The subscript $\text{m}$ denotes that a coordinate system has a metric scale, while the subscript $\text{f}$ denotes that the coordinate system has a freely chosen scale. In this illustration also the colours indicate whether units are metric (red) or not (blue). The algorithm optimizes world-to-camera poses $\mat{T}_{\text{C}_\text{f}\text{W}_\text{f}}$ which are directly used for the photometric energy. IMU factors use IMU-to-world poses $\mat{T}_{\text{W}_\text{m}\text{I}}$. The transformation $\mat{T}_{\text{W}_\text{m}\text{W}_\text{f}}$ between the metric and the non-metric world is another optimization variable, which does not only include scale, but also a rotation for gravity alignment. $\mat{T}_{\text{C}_\text{m}\text{C}_\text{f}}$ includes only scale, the same one as in $\mat{T}_{\text{W}_\text{m}\text{I}}$.  The IMU-to-camera transformation $\mat{T}_{\text{C}_\text{m}\text{I}}$ is known from calibration.}

\label{fig:systems}

\end{figure}

The coordinate systems we use in this paper are
\begin{description}
    \item[$\text{W}_\text{m}$] metric world,
    \item[$\text{W}_\text{f}$] world with freely chosen scale,
    \item[$\text{C}_\text{m}$] metric camera,
    \item[$\text{C}_\text{f}$] camera with freely chosen scale,
    \item[$\text{I}$] IMU (metric).
\end{description}
An overview of these systems including the transformations between them is also given in Fig.~\ref{fig:systems}. The reason why world and camera systems exist twice is that scale and gravity direction might not be observable from the beginning, hence the visual system starts estimating camera poses $\mat{T}_{\text{C}_\text{f}\text{W}_\text{f}}$  with a freely chosen scale, which can later be converted to a metric scale using the to-be-optimized variable $\mat{T}_{\text{W}_\text{m}\text{W}_\text{f}}$, which includes scale and rotation for gravity alignment. The transformation from non-metric to metric camera $\mat{T}_{\text{C}_\text{m}\text{C}_\text{f}}$ does not contain any additional variables, as it uses the scale of $\mat{T}_{\text{W}_\text{m}\text{W}_\text{f}}$, but identity rotation. Note also that IMU-to-world poses do not act as additional optimization variables, but are calculated from world-to-camera poses, which are optimized.

\section{MODEL}

In this section, we detail our formulation of a visual-inertial, rolling-shutter-aware energy, including brief reviews of previous methods that our method builds upon. The model described here only applies to the optimization backend, which jointly optimizes depth and keyframe variables, while the visual-inertial frontend that provides initializations operates as in \cite{von2018direct} and assumes global-shutter data. Hence, the words frame and keyframe are used interchangeably. As shown in Fig.~\ref{fig:systems}, there is a metric and a non-metric world. For the photometric energies, non-metric poses are used throughout. This has the advantage that tracking can be started right at the beginning of a sequence, while scale and gravity direction are optimized using the IMU information, instead of having to wait until scale is observable. The IMU factors, on the other hand, are calculated using metric poses.

\subsection{Photometric Energy}

To model the world-to-camera pose of frame $i$, i.e.\ the pose that converts a point in the world frame to the corresponding point in the camera frame, a constant twist model as in \cite{schubert2018direct} is used to parametrize the pose as a function of time,
\begin{align}
    \label{eq:posefun}
    \mat{T}_i(t) = \exp(\vecsym{\xi}_i t)\mat{T}_i^0\,,
\end{align}
where $\vecsym{\xi}_i\in\R^6$ is the twist for frame $i$ and $\mat{T}_i^0$ the central pose, corresponding to $t=0$. The algorithm is operating on pre-undistorted images, hence the capture time of a pixel at coordinates $(x,y)$ in the undistorted image is given by
\begin{align}
    \label{eq:time}
    t(x,y) = f_\text{d}(x,y) - y_0\,.
\end{align}
The distortion function $f_\text{d}$ is known from camera calibration. It maps the point into the original, distorted image, where only the $y$-coordinate is relevant for the capture time, as this is the readout direction of the rolling-shutter sensor. The offset $y_0$ is the vertical middle of the distorted image. We are free to measure time in pixel units, only later when the twist is compared to the IMU variables and measurements, the correct time conversion factor has to be chosen.

As in \cite{engel2018direct}, the photometric energy is a triple sum over the current set of keyframes $\mathcal{F}$, the points $\mathcal{P}_i$ hosted in a specific keyframe $i$ and the set of keyframes $\obs(\vec{p})$ in which a specific point $\vec{p}$ is observed,
\begin{align}
    E_\text{ph} = \sum_{i\in\mathcal{F}}\sum_{\vec{p}\in\mathcal{P}_i}\sum_{j\in\obs(\vec{p})} E_{\vec{p}j}\,.
\end{align}
The energy contribution $E_{\vec{p}j}$ by the observation of point $\vec{p}$ in frame $j$ is obtained by comparing intensities at the point's location in the host frame and its location when projected into the target frame. The projection into the target frame is not straightforward, as the pose of the target frame is needed, but the pose of the target frame depends on time, i.e. the pixel coordinate in the target frame, which in turn depends on the pose. To solve this mutual dependency, the rolling-shutter constraint \cite{meingast2005geometric} is solved iteratively as in \cite{schubert2018direct}.

\subsection{Visual-Inertial Factors}

We use IMU factors with preintegrated measurements as implemented in GTSAM\footnote{\url{https://bitbucket.org/gtborg/gtsam}} (based on \cite{forster2017manifold, carlone2014eliminating, lupton2012visual}), as they have been used in \cite{von2018direct, usenko2016direct}. Note that poses in this subsection are in the metric world, so in practice they have to be calculated from the non-metric camera poses using the current estimate for $\mat{T}_{\text{W}_\text{m}\text{W}_\text{f}}$. The state of frame $i$ consists of the pose (rotation $\mat{R}_i$ and translation $\vec{p}_i$), a translational velocity $\vec{v}_i$ of the IMU in the metric world frame and a bias vector $\vec{b}_i\in\R^6$,
\begin{align}
    \vec{s}_i=\left[\mat{R}_i, \vec{p}_i, \vec{v}_i, \vec{b}_i\right]\,.
\end{align}
From keyframe $i$ to the next keyframe $j$, measurements are integrated to obtain a prediction for the state of keyframe $j$. Starting with $\Delta\vec{p}=\vecsym{0}$, $\Delta\vec{v}=\vecsym{0}$ and identity rotation $\Delta\mat{R}=\mat{I}$, those quantities are iteratively updated as
\begin{align}
\label{eq:integr1}
    \Delta\vec{p} &\leftarrow \Delta\vec{p} + \Delta\vec{v}\Delta t\,,\\
    \Delta\vec{v} &\leftarrow \Delta\vec{v} + \Delta\mat{R}(\vec{a}-\vec{b}_i^\text{a})\Delta t\,,\\
    \Delta\mat{R} &\leftarrow\exp((\vecsym{\omega}-\vecsym{b}_i^\text{g})\Delta t)\,.
    \label{eq:integr3}
\end{align}
Here, $\Delta t$ is the time difference between two IMU measurements, $\vec{a}$ the current accelerometer measurement, $\vecsym{\omega}$ the current gyroscope measurement, $\vec{b}_i^\text{a}$ the three accelerometer components of the bias $\vec{b}_i$ corresponding to frame $i$ and $\vec{b}_i^\text{g}$ the three gyroscope components. In this case, the exponential maps 3D rotational velocities to rotation matrices in $\SO(3)$.

Integrating all measurements between keyframe $i$ and keyframe $j$ as in Eqs.~\ref{eq:integr1}-\ref{eq:integr3} yields the preintegrated measurements $\Delta\mat{R}_{ij}$, $\Delta\mat{v}_{ij}$ and $\Delta\mat{p}_{ij}$. To save computation time, the preintegration is not redone once the bias changes. Instead, the preintegrated measurements are linearized as functions of the bias. These linearized functions will be denoted $\Delta\mat{R}_{ij}(\vec{b}_i^\text{g})$, $\Delta\mat{v}_{ij}(\vec{b}_i^\text{g},\vec{b}_i^\text{a})$ and $\Delta\mat{p}_{ij}(\vec{b}_i^\text{g}, \vec{b}_i^\text{a})$ and are calculated as detailed in \cite{forster2017manifold}, Eq.~44.

The state predictions for frame $j$ are then calculated as
\begin{align}
\hat{\mat{R}}_j &= \mat{R}_i\Delta\mat{R}_{ij}(\vec{b}_i^\text{g})\,,\\
    \hat{\vec{p}}_j &= \vec{p}_i + (t_j-t_i)\vec{v}_i+\frac{1}{2}(t_j-t_i)^2\vec{g}
     +\mat{R}_i\Delta\mat{p}_{ij}(\vec{b}_i^\text{g}, \vec{b}_i^\text{a})\,,\\
    \hat{\vec{v}}_j &= \vec{v}_i + (t_j-t_i)\vec{g} + \mat{R}_i\Delta\mat{v}_{ij}(\vec{b}_i^\text{g},\vec{b}_i^\text{a})\,,
\end{align}
where $\vec{g}$ is the gravity vector and $t_i$ and $t_j$ are the timestamps of frames $i$ and $j$. 

The residuals are then calculated as
\begin{align}
    \vec{r}_{\Delta\mat{R}_{ij}}&=\log\left(\hat{\mat{R}}_j^\top\mat{R}_j\right)\,,\\
    \vec{r}_{\Delta\vec{v}_{ij}} &= \mat{R}_i^\top (\vec{v}_j - \hat{\vec{v}}_j)\,,\\
    \vec{r}_{\Delta\vec{p}_{ij}} &= \mat{R}_i^\top (\vec{p}_j - \hat{\vec{p}}_j)\,,\\
    \vec{r}_{\vec{b}_{ij}} &= \vec{b}_j - \vec{b}_i\,.
\end{align}
which are stacked in a residual vector $\vec{r}_{ij}$. These residuals then lead to the energy contribution
\begin{align}
    E_{ij} = \vec{r}_{ij}^\top \matsym{\Sigma} \vec{r}_{ij}\,,
\end{align}
with appropriate covariances $\matsym{\Sigma}$ as derived in \cite{forster2017manifold}. Summing these energies over the set of pairs of consecutive frames $\mathcal{C}$ yields the energy
\begin{align}
    E_\text{IMU} = \sum_{(i,j)\in\mathcal{C}} E_{ij}\,.
\end{align}

\begin{figure}[t]
    \centering
    \includegraphics[width=0.9\linewidth]{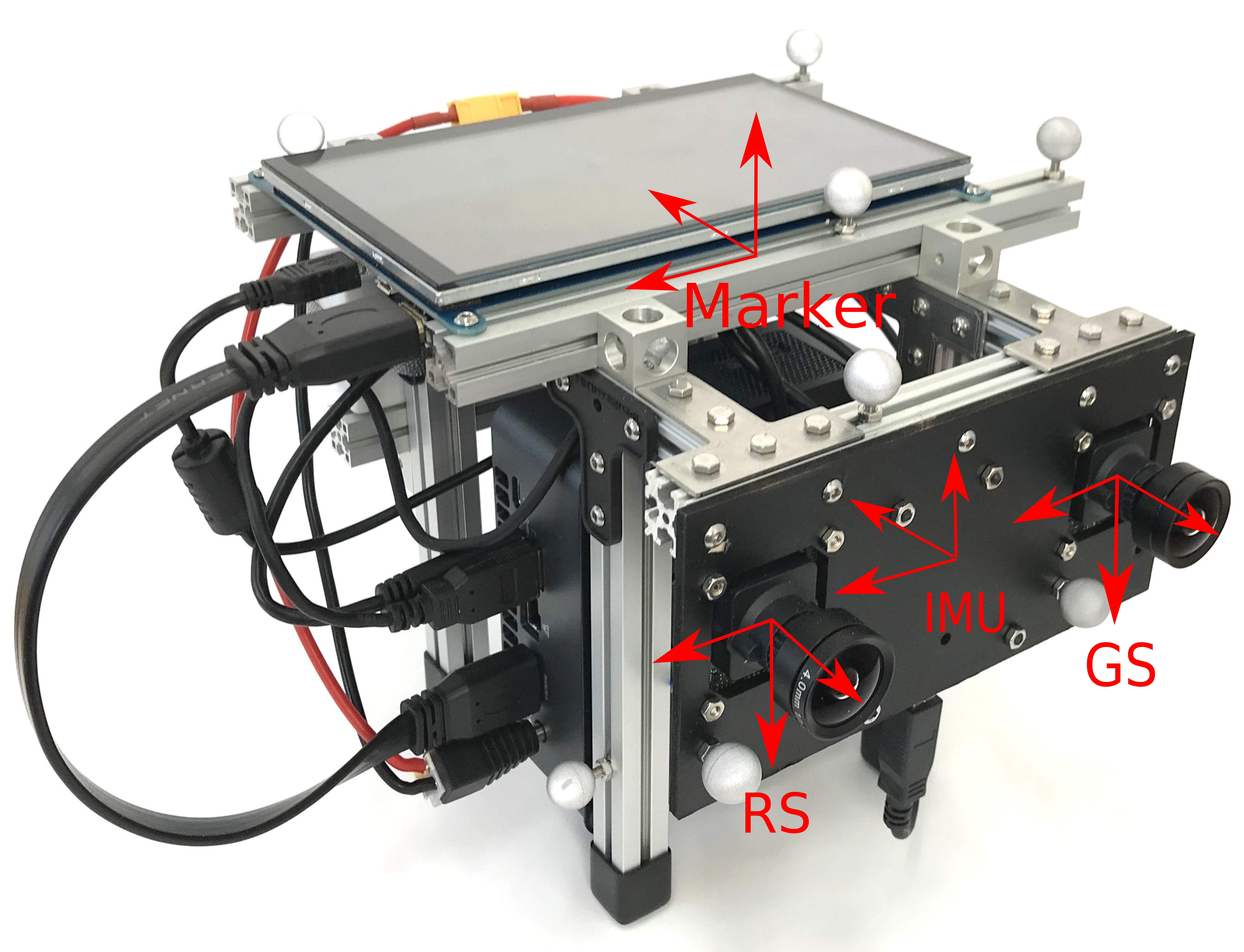}
    \caption{Camera setup that was used to acquire our new dataset. One camera is set to global-shutter mode and the other camera is set to rolling-shutter mode. Both cameras are hardware-synchronized with the IMU, and the transformations between all frames are pre-calibrated. Ground-truth data is recorded using a motion capture system. Time alignment for all sequences is done by aligning rotational velocities computed from the motion capture system and the gyroscope data.}
    \label{fig:camera}
\end{figure}

\begin{figure*}
    \centering
    \includegraphics[width=.95\linewidth]{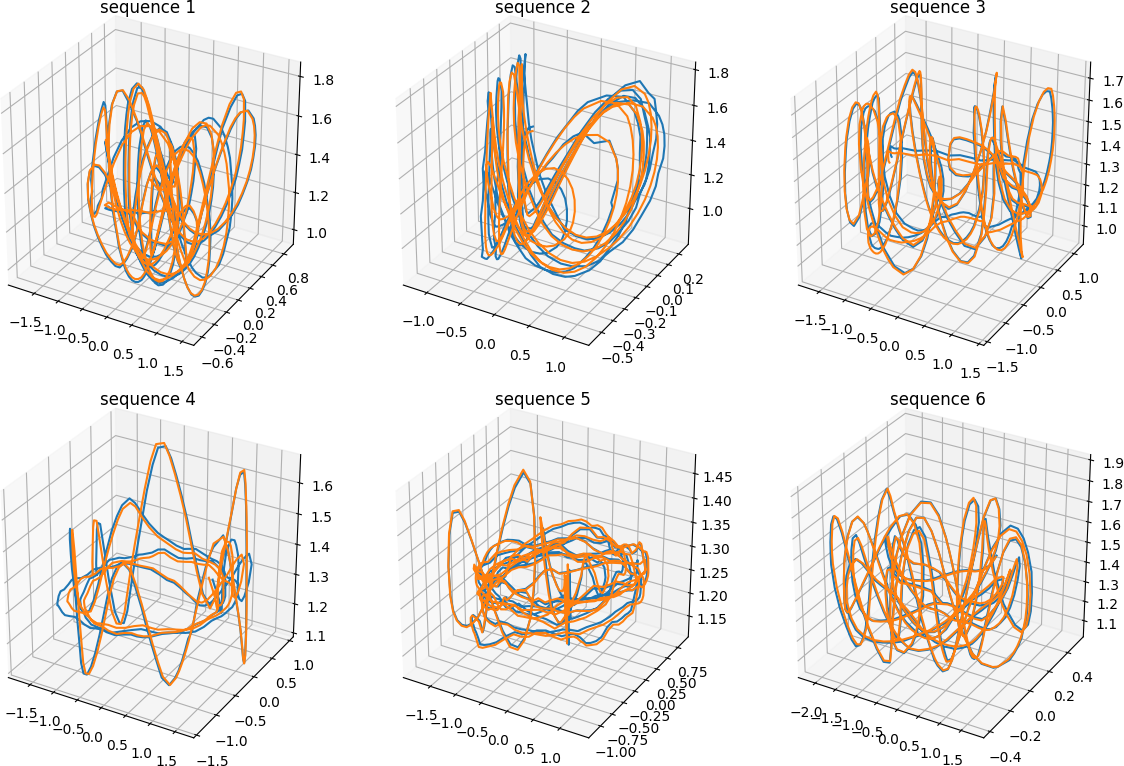}
    \caption{Trajectory plots for the first 6 sequences after $\SE(3)$ alignment of reconstructed trajectories (blue) and ground truth (orange), axes in meters. To give representative examples, a run with error $e_\text{ate}$ close to the median error is shown for each sequence.}
    \label{fig:traj_plots}
\end{figure*}

\subsection{Twist Energy}
In \cite{schubert2018direct}, the system is stabilized by a prior for the twist $\vecsym{\xi}_i$ introduced in Eq.~\ref{eq:posefun}, that assumes a smooth motion between keyframes. Using an IMU, we are in a much more comfortable situation, as it provides high-frequency measurements and hence a much more up-to-date prior for the twist, leading to the novel formulation for the twist energy that we propose. From the IMU, we directly obtain a gyroscope measurement $\vecsym{\omega}$, which is biased by $\vec{b}^\text{g}$, and the state includes the translational velocity $\vec{v}$ as an optimization variable. For the ease of notation, we drop the keyframe index $i$, but still all variables belong to a certain keyframe, in particular to the timestamp of its mid-pose $\mat{T}_i^0$.

The velocity $\vec{v}$ is the velocity of the IMU in the metric world frame $\text{W}_\text{m}$. We first rotate this velocity into the IMU frame,
\begin{align}
\vec{v}^\text{IMU} &= \mat{R}_{\text{I}\text{W}_\text{m}} \vec{v}\\
&=\mat{R}_{\text{C}_\text{m}\text{I}}^{-1} \mat{R}_{\text{C}_\text{m}\text{C}_\text{f}} \mat{R}_{\text{C}_\text{f}\text{W}_\text{f}}  \mat{R}_{\text{W}_\text{m}\text{W}_\text{f}}^{-1}   \vec{v}\,,
\end{align}
where $\mat{R}_{BA}$ is the rotation of $\mat{T}_{BA}$, thus trivially $\mat{R}_{\text{C}_\text{m}\text{C}_\text{f}}=\mat{I}$. We cannot use $\mat{T}_{\text{W}_\text{m}\text{I}}^{-1}$ directly to obtain $\mat{R}_{\text{I}\text{W}_\text{m}}$, as $\mat{T}_{\text{W}_\text{m}\text{I}}$ is not an optimization variable. It needs to be expressed as a function of the non-metric camera poses $\mat{T}_{\text{C}_\text{f}\text{W}_\text{f}}$.

Now we have the required quantities for the IMU twist,
\begin{align}
    \vecsym{\xi}^\text{IMU} = \begin{bmatrix}
        \vec{v}^\text{IMU}\\
        \vecsym{\omega} - \vec{b}^\text{g}
    \end{bmatrix}\,,
\end{align}
which could be used to approximate the motion of the IMU. We are, however, interested in the motion of the camera as given in Eq.~\ref{eq:posefun}, so the twist has to be converted using the adjoint $\Adj(\mat{T}_{\text{C}_\text{m}\text{I}})$ of the relative pose between camera und IMU,
\begin{align}
    \vecsym{\xi}^{\text{cam}} = -\Adj(\mat{T}_{\text{C}_\text{m}\text{I}})\vecsym{\xi}^\text{IMU}\,.
\end{align}
The adjoint is a $6\times6$ matrix and has the property
\begin{align}
    \mat{T}\exp(\vecsym{\delta}) = \exp\left(\Adj(\mat{T})\vecsym{\delta}\right)\mat{T}\,.
\end{align}
Thus, we can show that acting with $\vecsym{\xi}^\text{IMU}$ on the IMU pose has the same effect as acting with $\vecsym{\xi}^{\text{cam}}$ on the camera pose:
\begin{align}
    &\mat{T}_{\text{W}_\text{m}\text{I}} \exp\left(\vecsym{\xi}^\text{IMU}t\right)\\
     &= \mat{T}_{\text{C}_\text{m}\text{W}_\text{m}}^{-1}\mat{T}_{\text{C}_\text{m}\text{I}}\exp(\vecsym{\xi}^\text{IMU}t)\\
    &= \mat{T}_{\text{C}_\text{m}\text{W}_\text{m}}^{-1}\exp\left(\Adj(\mat{T}_{\text{C}_\text{m}\text{I}})\vecsym{\xi}^\text{IMU}t\right)\mat{T}_{\text{C}_\text{m}\text{I}}\\
    &= \left(\exp\left(-\Adj(\mat{T}_{\text{C}_\text{m}\text{I}})\vecsym{\xi}^\text{IMU}t\right)\mat{T}_{\text{C}_\text{m}\text{W}_\text{m}}\right)^{-1}\mat{T}_{\text{C}_\text{m}\text{I}}\\
    &= \left(\exp\left(\vecsym{\xi}^\text{cam}t\right)\mat{T}_{\text{C}_\text{m}\text{W}_\text{m}}\right)^{-1}\mat{T}_{\text{C}_\text{m}\text{I}}\,.
\end{align}

So far the translational components are in metric units. To obtain an appropriate prior for the twist in Eq.~\ref{eq:posefun}, the translational components $\vecsym{\xi}^\text{cam}_\text{t}$ have to be divided by the scale $s$ of the transformation $\mat{T}_{\text{W}_\text{m}\text{W}_\text{f}}$, while the rotational components $\vecsym{\xi}^\text{cam}_\text{r}$ stay unaffected. Hence, the prior for the twist is
\begin{align}
    \tilde{\vecsym{\xi}} = t_\text{d}\begin{bmatrix}
        \vecsym{\xi}^\text{cam}_\text{t}/s\\
        \vecsym{\xi}^\text{cam}_\text{r}
    \end{bmatrix}\,,
\end{align}
which is also scaled by the time difference $t_\text{d}$ between two consecutive image rows, as this is the unit of time chosen in Eq.~\ref{eq:time}.
Finally, the energy contribution is a weighted squared deviation, summed over all frames:
\begin{align}
    E_\text{twist} = \sum_\mathcal{F}(\tilde{\vecsym{\xi}}-\vecsym{\xi})^\top\matsym{\Sigma}(\tilde{\vecsym{\xi}}-\vecsym{\xi})\,,
\end{align}
with a manually chosen diagonal matrix $\matsym{\Sigma}$.

\subsection{Optimization}

We optimize the total energy
\begin{align}
    E = E_\text{ph} + \alpha E_\text{IMU} + \beta E_\text{twist}
\end{align}
using Gauss-Newton optimization. The parameters $\alpha$ and $\beta$ are balancing weights. As in \cite{engel2018direct}, keyframes are marginalized once the set of keyframes grows too large. Once a variable is part of the marginalization term, its linearization point is not changed any more to keep the system consistent. This is usually a good approximation, as the state estimates do not fluctuate heavily, but for scale, this is not the case. We therefore use the approach of \cite{von2018direct} and keep a second version of the marginalization Hessian that only contains newer IMU factors, which can be used once the scale estimate moves too far from the linearization point of the current marginalization Hessian. As the newly introduced twist energy depends on scale, we also include it with two different linearization points in the two versions of the marginalization Hessian.

\section{NEW ROLLING-SHUTTER DATASET}

Since we want to compare our novel rolling-shutter VIO method to the baseline global-shutter method not only on rolling-shutter images, but also on global-shutter images, we need a dataset that provides both simultaneously. 
To the best of our knowledge, there is no such dataset suitable for VIO evaluation. Therefore, we recorded our own dataset and make it publicly available. 
This dataset comprises 10 challenging indoor sequences spanning a total of around \SI{7}{\minute} and \SI{475}{\meter} traversed distance. Tab.~\ref{tab:sequences} shows some statistics of the individual sequences, where the mean velocities and accelerations are computed as numerical derivatives of the ground-truth IMU poses.

\begin{table}
    \caption{Sequences in our rolling-shutter dataset}
    \label{tab:sequences}
    \centering
    \begin{tabular}{lrrrrrr}
        \toprule
        Seq. & Duration & Length & \multicolumn{2}{c}{Mean Velocity} & \multicolumn{2}{c}{Mean Acceleration} \\
        & \multicolumn{1}{c}{[\si{\second}]} & \multicolumn{1}{c}{[\si{\meter}]} & \multicolumn{1}{c}{[\si{\meter/\second}]} & \multicolumn{1}{c}{[\si{\degree/\second}]} & \multicolumn{1}{c}{[\si{\meter/\second\squared}]} & \multicolumn{1}{c}{[\si{\degree/\second\squared}]}  \\
        \midrule
        1 & 40 & 46 & 1.1 & 62 & 3.2 & 233 \\
        2 & 27 & 37 & 1.4 & 73 & 4.1 & 271 \\
        3 & 50 & 44 & 0.9 & 56 & 2.3 & 220 \\
        4 & 38 & 30 & 0.8 & 39 & 1.1 & 148 \\
        5 & 85 & 57 & 0.7 & 43 & 0.8 & 167 \\
        6 & 43 & 51 & 1.2 & 50 & 2.6 & 252 \\
        7 & 39 & 45 & 1.1 & 92 & 2.7 & 148 \\
        8 & 53 & 46 & 0.9 & 91 & 1.8 & 103 \\
        9 & 45 & 46 & 1.0 & 137 & 3.8 & 539 \\
        10 & 54 & 41 & 0.7 & 116 & 2.1 & 605 \\
        \midrule
        Total & 475 & 442 & 0.9 & 75 & 2.2 & 267 \\
        \bottomrule
    \end{tabular}
\end{table}

The sensor setup as depicted in Fig.~\ref{fig:camera} includes two uEye UI-3241LE-M-GL cameras by IDS with Lensagon BM4018S118 lenses by Lensation. 
The cameras record time-synchronized images at \SI{20}{\hertz} with the left camera running in global-shutter mode and the right camera in rolling-shutter mode and a row time difference of approximately \SI{29.47}{\micro\second}.
The \num{1280}x\num{1024} grayscale images are captured with a linear response function at \SI{16}{bit} to retain the full dynamic range of the imager, and in our direct VIO approach we additionally use pre-calibrated vignette compensation as in \cite{engel2018direct}.
The Bosch BMI160 accelerometer and gyroscope readings at \SI{200}{\hertz} are time-synchronized with the cameras in hardware.
Ground-truth motion is recorded with an OptiTrack Flex13 motion capture system that uses ceiling-mounted cameras to track IR-reflective markers mounted on the sensor setup at \SI{120}{\hertz}. A simple median filter discards outlier motion capture poses and linear interpolation is used to compute reference poses at keyframe times.

Similar to \cite{schubert2018tum} we calibrate camera and IMU intrinsics, as well as extrinsics between both cameras, the IMU and the motion capture markers. In all calibration sequences both cameras are in global-shutter mode to ensure accurate results.
Since the motion capture poses are not time-synchronized during recording, we perform temporal alignment for each evaluation sequence by aligning gyroscope measurements to angular velocities estimated from motion capture.

With the dataset we provide our calibration results, pre-processed sequences with IMU intrinsics compensated (scale, axis-alignment, constant bias) and ground-truth poses geometrically and temporally aligned to the IMU frame. On top of that, raw data and calibration sequences are also available to facilitate custom calibration schemes.

\section{QUANTITATIVE EVALUATION}

\begin{figure}
    \centering
    \includegraphics[width=0.9\linewidth]{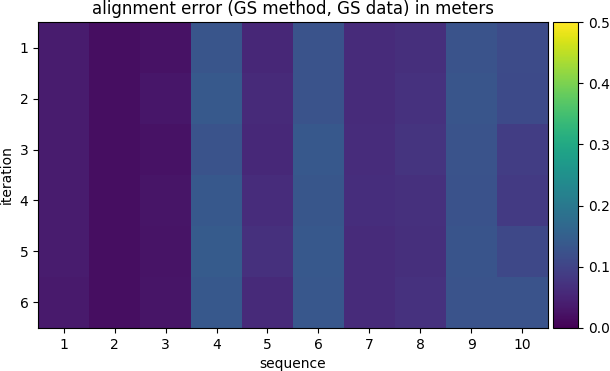}
    \\[0.2cm]
    \includegraphics[width=0.9\linewidth]{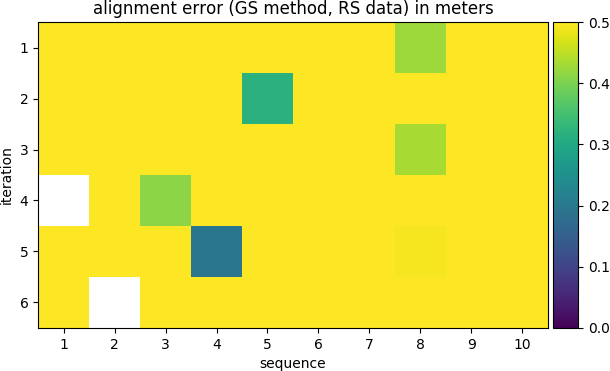}
    \\[0.2cm]
    \includegraphics[width=0.9\linewidth]{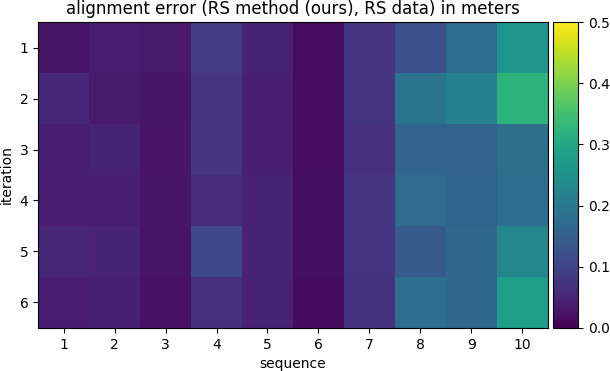}
    \caption{Colour-coded absolute trajectory error $e_\text{ate}$ after $\SE(3)$ alignment. Each of the 10 sequences from our new dataset has been run 6 times in three different modes: the global-shutter baseline method operating on global-shutter images (top); the global-shutter baseline method operating on rolling-shutter images (middle); our new rolling-shutter method operating on rolling-shutter images. White squares correspond to runs that failed entirely due to numerical instability. The baseline method produces stable results on global-shutter data, but mostly fails on rolling-shutter data. With our new method, similar results as for the baseline method on global-shutter data can be achieved.
    }
    \label{fig:colorplots}
\end{figure}

We run extensive evaluations on our newly taken dataset. The baseline method is a global-shutter method as in \cite{von2018direct}. It also integrates visual and inertial measurements, but does not feature a rolling-shutter model. Both \cite{von2018direct} and our new method are operated with \num{2000} points and a maximum of 7 keyframes. As the new dataset contains pairs of similar rolling-shutter and global-shutter sequences, not only can we compare the performance of the new method with the baseline method operating on rolling-shutter data, but also with the baseline method operating on global-shutter data.

Inertial methods can observe scale even with a monocular camera, provided a non-degenerate motion. In our case, the final scale estimate is used to convert the optimized non-metric poses to metric poses. Thus, to compare the estimated trajectories with the corresponding ground truth, no scale alignment has to be performed. The evaluation metric in this section is the absolute trajectory error after $\SE(3)$ alignment, defined as
\begin{align}
    e_\text{ate} = \min_{\mat{T}\in \SE(3)} \sqrt{\frac{1}{n}\sum_i \left\|\mat{T}\vec{p}_i-\hat{\vec{p}}_i\right\|^2}\,,
\end{align}
where $i$ is iterated over all keyframes in the whole sequence, and $n$ is the number of such keyframes. The keyframe positions estimated by an algorithm are denoted $\vec{p}_i$ and the corresponding ground-truth positions are denoted $\hat{\vec{p}}_i$.

To gather statistics, each sequence of the dataset has been run 6 times for each mode, with different random seeds which influences point selection. The three modes are 
\begin{itemize}
\item running the global-shutter baseline method on global-shutter images;
\item running the global-shutter baseline method on rolling-shutter images;
\item running our new rolling-shutter method on rolling-shutter images.
\end{itemize}
We do not compare with the similar purely visual method in~\cite{schubert2018direct}, as it cannot observe scale, which makes a fair comparison difficult.
In Fig.~\ref{fig:colorplots} the results of these experiments are shown in a colour plot. Each coloured square corresponds to one run and 6 squares in the same column correspond to the 6 runs of the respective sequence. The colour of the square encodes the absolute trajectory error $e_\text{ate}$. The results of the global-shutter method on global-shutter data show a stable and accurate performance. Running the same algorithm on rolling-shutter data drastically changes the results. Apart from five runs, all runs are beyond the colour scale. Individual inspection showed that this is not inaccuracy, but instability, as the tracking results diverged far from the ground truth. Two runs failed entirely, which means the system became numerically unstable. These results are interesting when compared to the results in~\cite{schubert2018direct} (a similar, purely visual method), where the global-shutter method was not entirely unstable, but often estimated trajectories with drift. Possibly, the rolling-shutter images can be approximately explained with a slightly altered trajectory, but if there is inertial data that contradicts this slightly altered trajectory, the system breaks.

Using the new rolling-shutter method redeems most of the problems with rolling-shutter data. Apart from sequence~10, it shows very stable performance with accuracies similar to the global-shutter method on global-shutter data.

A more quantitative comparison is given in Tab.~\ref{tab:ate}. For each of the three modes and for each of the 10 sequences, the median of the absolute trajectory error $e_\text{ate}$ over all 6 runs is given. The errors of the global-shutter method on rolling-shutter data are larger than for the other two modes by orders of magnitude in most cases. A comparison of the global-shutter method on global-shutter data with the rolling-shutter method on rolling-shutter data does not yield a clear preference for all sequences. There are more sequences with more accurate results for the global-shutter method on global-shutter data, but also some sequences where the rolling-shutter method on rolling-shutter data is more accurate. One possible explanation for the latter case is that due to the time shift between rows, rolling-shutter images provide additional information about velocities. On the other hand, one reason for less accurate results by the rolling-shutter method might be the constant-twist assumption, which is violated in the presence of large accelerations. This reasoning is supported by the fact that sequence~10, the sequence with the largest mean rotational acceleration, is the sequence with the least accurate results. Remedy for the constant-twist model would be brought by a shorter row time difference, as then the twist only has to be valid over a shorter time interval. The rolling shutter in our dataset is rather at the slower end of the scale, so the shutter may well be faster in other use cases.

A visual impression of the tracking result of our algorithm on the first 6 sequences is given in Fig.~\ref{fig:traj_plots}. It shows the estimated camera trajectory together with ground truth, after $\SE(3)$ alignment. For each sequence, a run with error $e_\text{ate}$ close to the median error was selected. Qualitatively, the estimated trajectories have very similar shapes compared to the ground truth, with small deviations visible.

One significant drawback of our method is runtime. As our approach combines energy terms and variables of \cite{schubert2018direct} and \cite{von2018direct}, it is slower than the sub-realtime performance reported in \cite{schubert2018direct}, so possible speedups remain an open challenge.

\begin{table}
\caption{Absolute trajectory error after $\SE(3)$ alignment, median over 6 runs, in meters. Minimum for each sequence in bold.}
\label{tab:ate}
\centering
\begin{tabular}{cccc}
\toprule
Seq. & GS method, GS data & GS method, RS data & Ours, RS data \\
\midrule
1 & \textbf{0.038} & \num{79.591} & \num{0.040} \\
2 & \textbf{0.018} & \num{40.725} & \num{0.044} \\
3 & \textbf{0.027} & \num{1.803} & \num{0.028} \\
4 & \num{0.137} & \num{0.970} & \textbf{0.079} \\
5 & \num{0.060} & \num{0.683} & \textbf{0.049} \\
6 & \num{0.135} & \num{2.352} & \textbf{0.017} \\
7 & \textbf{0.061} & \num{28.336} & \num{0.075} \\
8 & \textbf{0.070} & \num{0.501} & \num{0.168} \\
9 & \textbf{0.128} & \num{218.152} & \num{0.168} \\
10 & \textbf{0.111} & \num{482.021} & \num{0.246} \\
\bottomrule
\end{tabular}
\end{table}

\section{Conclusion}

In this paper, we present a direct sparse visual-inertial odometry approach for rolling-shutter cameras. We estimate poses, linear velocities and biases for a set of keyframes. These variables are coupled to the twist used to represent the continuous motion of the camera that is needed to model rolling-shutter projection. This way, unlike global-shutter methods, we estimate velocities and biases from both IMU measurements and rolling-shutter effects.

Due to the lack of suitable datasets, we recorded a new dataset with global-shutter and rolling-shutter images, IMU data and ground-truth poses from motion capture and make it publicly available. 
Our evaluation on this dataset shows that we can achieve similar accuracy to conventional methods on global-shutter data by explicitly modelling and exploiting rolling-shutter effects in the visual-inertial odometry.

\addtolength{\textheight}{-12cm}   







\bibliographystyle{IEEEtran}
\bibliography{IEEEabrv,IEEEexample}

\end{document}